\documentclass{article}

\usepackage[final]{neurips_2025}

\usepackage{booktabs}      
\usepackage{colortbl}      
\usepackage[table]{xcolor} 
\usepackage{arydshln}
\usepackage{pifont}
\usepackage{enumitem}
\usepackage{comment}
\usepackage{colortbl, xcolor, graphicx, booktabs}
\usepackage{pgf}
\usepackage{multirow}

\usepackage{graphicx}
\usepackage[normalem]{ulem}
\usepackage{xcolor, colortbl}
\usepackage[utf8]{inputenc} 
\usepackage[T1]{fontenc}    
\usepackage{hyperref}       
\usepackage{url}            
\usepackage{booktabs}       
\usepackage{amsfonts}       
\makeatletter
\providecommand{\@trackname}{}
\makeatother
\usepackage{nicefrac}       
\usepackage{microtype}      
\usepackage{xcolor}         

\usepackage{graphicx}       
\usepackage{epstopdf}       
\DeclareGraphicsExtensions{.pdf,.png,.jpg}

\graphicspath{{./}{NeurIPS25/Styles/sec/img/}}  

\title{\includegraphics[width=\textwidth]{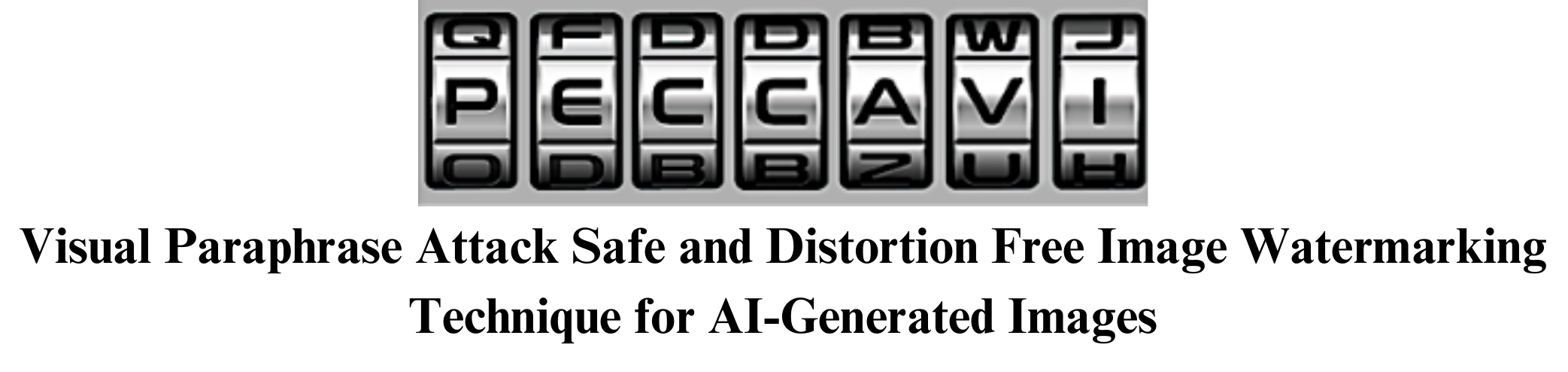}}

\author{
    \textbf{Shreyas Dixit}\textsuperscript{1}\thanks{Equal contribution.} \quad
    \textbf{Ashhar Aziz}\textsuperscript{2}\footnotemark[1] \quad
    \textbf{Shashwat Bajpai}\textsuperscript{3}\footnotemark[1] \\
    \textbf{Vasu Sharma}\textsuperscript{4} \quad
    \textbf{Aman Chadha}\textsuperscript{5,6}\thanks{Work does not relate to position at Amazon.} \quad
    \textbf{Vinija Jain}\textsuperscript{5} \quad
    \textbf{Amitava Das}\textsuperscript{7} \\ \\
    \textsuperscript{1}VIIT Pune, India \quad
    \textsuperscript{2}IIIT Delhi, India \quad
    \textsuperscript{3}BITS Pilani Hyderabad, India \\
    \textsuperscript{4}Meta AI, USA \quad
    \textsuperscript{5}Stanford University, USA \quad
    \textsuperscript{6}Amazon GenAI, USA \\
    \textsuperscript{7}AI Institute, University of South Carolina, USA
}

\begin{document}

\maketitle

\begin{abstract}
A report by the European Union Law Enforcement Agency predicts that by 2026, up to 90\% of online content could be synthetically generated \cite{EUROPOL}, raising concerns among policymakers, who cautioned that "\uline{Generative AI could act as a force multiplier for political disinformation. The combined effect of generative text, images, videos, and audio may surpass the influence of any single modality}" \cite{janjevagenai2023}. In response, California's Bill AB 3211 mandates the watermarking \cite{california_act} of AI-generated images, videos, and audio. However, concerns remain regarding the vulnerability of invisible watermarking techniques to tampering and the potential for malicious actors to bypass them entirely. Generative AI-powered de-watermarking attacks, especially the newly introduced visual paraphrase attack \cite{barman2024brittlenessaigeneratedimagewatermarking}, have shown an ability to fully remove watermarks, resulting in a paraphrase of the original image. This paper introduces PECCAVI, the first visual paraphrase attack safe and distortion free
image watermarking technique. In visual paraphrase attacks, an image is altered while preserving its core semantic regions, termed Non-Melting Points (NMPs). PECCAVI strategically embeds watermarks within these NMPs and employs multi-channel frequency domain watermarking. It also incorporates noisy burnishing to counter reverse-engineering efforts aimed at locating NMPs to disrupt the embedded watermark, thereby enhancing durability. PECCAVI is model-agnostic. All relevant resources and codes will be open-sourced\footnote{PECCAVI is protected under a USA patent. In accordance with the Patent Act, adoption or research involving this technology for non-profit purposes is strictly prohibited.}.    
\end{abstract}
\vspace{-4mm}
\section{Introduction - the Necessity \& Urgency}

\begin{figure*}[t!]
  \centering
  \includegraphics[width=\textwidth]{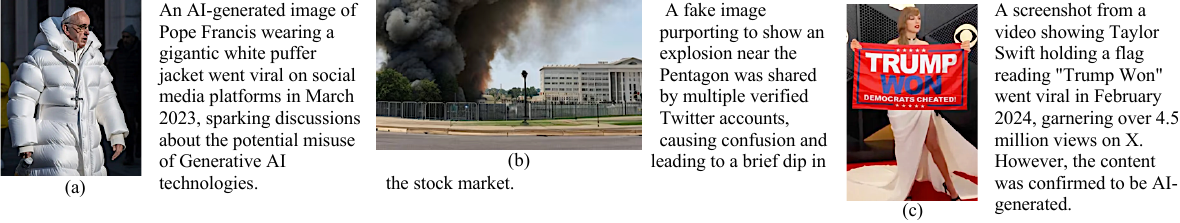}
  \vspace{-6mm}
  \caption{Some examples of the negative impacts of AI-generated images: (a) \textit{``That Viral Image Of Pope Francis Wearing A White Puffer Coat Is Totally Fake"} 
\href{https://www.forbes.com/sites/mattnovak/2023/03/26/that-viral-image-of-pope-francis-wearing-a-white-puffer-coat-is-totally-fake/}{(Forbes story)}, (b) \textit{'Verified' Twitter accounts share fake image of 'explosion' near Pentagon, causing confusion} \href{https://www.cnn.com/2023/05/22/tech/twitter-fake-image-pentagon-explosion/index.html}{(CNN's cover story)}, and (c) \textit{Viral Video Of Taylor Swift 'Endorsing' Donald Trump Is Completely Fake} \href{https://www.forbes.com/sites/mattnovak/2024/02/05/viral-video-of-taylor-swift-endorsing-donald-trump-is-completely-fake/}{(Forbes story)}.
  }
  \label{fig:ai_misinformation}
  \vspace{-3mm}
\end{figure*}

The proliferation of text-to-image generative AI models like Stable Diffusion(s) \citep{Rombach_2022_CVPR}, DALL-E(s) \citep{ramesh2021zeroshottexttoimagegeneration,ramesh2022hierarchicaltextconditionalimagegeneration,BetkerImprovingIG}, Midjourney \citep{Midjourney2024}, and Imagen \citep{saharia2022photorealistictexttoimagediffusionmodels} has revolutionized visual content creation, unlocking unprecedented creative potential. However, this rapid evolution and widespread accessibility presents significant challenges, particularly concerning the misuse of AI-generated images.

In March 2023, an open letter \citep{aihalt2023} signed by numerous AI experts and industry leaders called for a six-month halt on the development of AI systems more advanced than GPT-4. The central concern noted in the letter \citep{aihalt2023} is ``\textit{Should we let machines flood our information channels with propaganda and untruth?}". While individual viewpoints on the notion of a moratorium may vary, the raised concern cannot be ignored. The findings of the latest (7\textsuperscript{th}) evaluation of the European Commission's Code of Conduct \citep{EU_code_of_conduct_2022} that seeks the eradication of mis/dis-information online reveals a decline in companies' responsiveness. The percentage of notifications reviewed by companies within 24 hours decreased, falling from 90.4\% in 2020 to 64.4\% in 2022. 

This decline likely reflects the increased accessibility of Gen AI models, leading to a notable influx of AI-generated content on the web. Approximately 3.2 billion images and 720,000 hours of video are uploaded to social media platforms daily \citep{Thomson2020} (as of 2020). With all existing image watermarking techniques proving brittle against Gen AI-powered de-watermarking attacks \citep{barman2024brittlenessaigeneratedimagewatermarking}, the need for developing robust, attack-resistant watermarking methods is more critical than ever.

AI-generated misinformation stands as one of the most formidable challenges in advancing responsible AI for society, as emphasized by leading figures including Geoffrey Hinton \citep{hinton_misinformation}, Bill Gates \citep{gates_misinformation}, and Sundar Pichai \citep{pichai_misinformation}, among others. Figure \ref{fig:ai_misinformation} illustrates recent instances of AI-generated misinformation that have caused significant disruptions.
\vspace{-2mm}
\section{Dewatermarking Attacks - Related Works} 
\label{sec:approach}
Digital watermarking has been a focus in computer vision research for 3–4 decades, primarily divided into two categories : (i) static, learning-free methods: such as DwtDctSVD \citep{navas2008dwt}, IA-DCT \citep{podilchuk1998image}, and IA-W \citep{podilchuk1998image}, among others; and (ii) learning-based methods, which represent the more contemporary, state-of-the-art approaches such as Stable Signature \citep{fernandez2023stablesignaturerootingwatermarks}, Tree-Ring Watermark \citep{wen2023treerings}, Watermark Anything Model (WAM) \citep{sander2024watermarklocalizedmessages}, and ZoDiac \citep{zhou2024zodiaccardiologistlevelllmframework} etc. In addition to developing and evaluating various watermarking methods, researchers have also explored de-watermarking techniques, including classical image-altering methods such as (a) \emph{brightness adjustment} \citep{brightness_adjustment}, (b) \emph{JPEG compression} \citep{jpeg_compression}, and (c) \emph{Gaussian noise addition} \citep{cox1997secure}. More recently, advanced Generative AI-powered techniques have been introduced, such as (d) regeneration attacks \citep{zhao2023invisible} and (f) adversarial purification \citep{pmlr-v162-nie22a}. 

While watermarking originated in computer vision, advancements in large language models (LLMs) have spurred interest in text watermarking. OpenAI, for instance, hinted at watermarking techniques for ChatGPT \citep{businessstandard2024}. Early LLM watermarking models by \citep{Kirchenbauer2023AWF} faced criticism after studies by \citep{sadasivan2024aigeneratedtextreliablydetected} and \citep{chakraborty-etal-2023-counter} showed that paraphrasing could effectively remove these watermarks. This has spurred interest in visual paraphrase attacks \citep{barman2024brittlenessaigeneratedimagewatermarking} on image watermarks, a technique enabled by advances in text-to-image and image-to-text systems.

\subsection{Visual Paraphrase attack}
The concept of \emph{visual paraphrasing attack}, first introduced in \citep{barman2024brittlenessaigeneratedimagewatermarking}, refers to generating variations of an image that retain the same semantic content while altering visual presentation. An illustration is visible in Figure \ref{fig:paraphrasing}, taken from the original paper. Unlike linguistic paraphrasing in natural language processing (e.g., “\emph{What is your age?}” vs. “\emph{How old are you?}”), visual paraphrasing utilizes image-to-image diffusion \citep{Rombach_2022_CVPR} system, to adjust an image’s visual representation while preserving its meaning. The authors examine two primary parameters for creating effective visual paraphrases: Strength and Guidance Scale. Their findings suggest that a specific range of Strength, paired with an optimal range of Guidance Scale, yields satisfactory dewatermarked paraphrased images. An example can be found in Figure 2. For a more detailed explanation, please refer to \citep{barman2024brittlenessaigeneratedimagewatermarking}. Further details are reported in the Appendix: Visual Paraphrase.

The visual paraphrase attack is comparable to other generative AI-powered attacks, such as regeneration attacks \citep{zhao2023invisible} and adversarial purification \citep{pmlr-v162-nie22a}. Although the recently proposed ZoDiac \citep{zhou2024zodiaccardiologistlevelllmframework} has demonstrated resilience against regeneration and adversarial purification attacks, the visual paraphrase attack remains the most effective method for completely removing watermarks.

The visual paraphrasing attack, introduced by \citep{barman2024brittlenessaigeneratedimagewatermarking}, takes a watermarked image and uses an image-to-image diffusion system to produce a watermark-free visual paraphrase. An alternative approach could involve: $ watermarked\;image\Rightarrow image\;captioning\;system \Rightarrow image\;caption \Rightarrow text\;to\;image\;system \Rightarrow watermark\;free\;image$, which we term \emph{open-ended visual paraphrase}. However, generating a near-identical image this way is nearly impossible due to the challenge of crafting a precise text prompt, resulting in countless variations and making the exact reproduction improbable, with potential seed values reaching $2^{64}$ \citep{pytorch_max_seed_discussion}. Additionally, text-to-image (T2I) systems are inherently stochastic, producing different outputs each time, even with the same input prompt. Our image-to-image approach, however, ensures reliable adherence to the original image’s appearance and meaning, even within this variable parameter space, delivering a consistent and structurally faithful visual paraphrase. Further discussion is in Appendix: Visual Paraphrase.

\section{PECCAVI: Visual Paraphrase Attack Safe \& Distortion-Free Image Watermarking Technique}

To the best of our knowledge, there is no existing work on visual paraphrase attack-safe watermarking, making it challenging to directly compare PECCAVI with similar techniques. In designing PECCAVI, we considered several fundamental questions: (i) \emph{where to place the watermark}, (ii) \emph{which watermarking technique to use}, (iii) \emph{the need for a more sophisticated detection mechanism}, (iv) \emph{how to assess resistance to visual paraphrase attacks}, and (v) \emph{whether the watermarking process distorts the original concept excessively}. We describe PECCAVI in detail with these guiding questions in mind to aid the reader’s understanding. The overall pipeline of PECCAVI is illustrated in Figure \ref{fig:peccavi_architecture}.

\subsection{Where to add watermark? - Non-Melting Points (NMPs)}
{Visual paraphrasing creates alternate visual representations of an image while preserving its core meaning. The key concept here is to identify regions that remain largely unaffected by paraphrasing. These relatively stable regions are ideal for embedding watermark signals, as they are less likely to be altered. We refer to these areas as Non-Melting Points/Regions (NMPs). Detecting NMPs involves two main steps:

\textbf{Saliency detection}: Salient region detection in image processing identifies the most ``\emph{salient}" or visually prominent areas within an image, based on unique features like color contrast, texture, or edges. Saliency detection is a well-established sub-discipline, with various methods; empirically, we found that \citep{9008576} XRAI performed best in our experiments, followed by MSI-Net \citep{kroner2020contextual} and  
hence we will be using these for our experiments in the rest of the paper. Refer to Figure \ref{fig:saliency_detection} for a detailed illustration.

\textbf{Non-Melting Points (NMPs)}: For a given image, we generate five automatically paraphrased versions using the method described in \citep{barman2024brittlenessaigeneratedimagewatermarking}. In each paraphrased image, we identify key regions and use Intersection over Union (IoU) to find the most stable areas—regions that consistently appear in similar locations across variations. These stable areas are referred to as Non-Melting Points (NMPs).
To refine our NMP selection, we apply Non-Maximum Suppression (NMS), which eliminates redundant overlapping boxes and retains only the most representative regions. The final set of NMPs is then mapped onto a predefined patch grid, where each patch is evaluated for stability.
Each region receives a stability score, which reflects how frequently it appears across the paraphrased images. Lower scores indicate greater consistency across variations, making those regions more reliable as NMPs. If no sufficiently stable regions are found, we include a default box to ensure robustness.

\vspace{-3mm}
\begin{figure}[ht!]
    \centering
    \includegraphics[width=1.0\columnwidth]{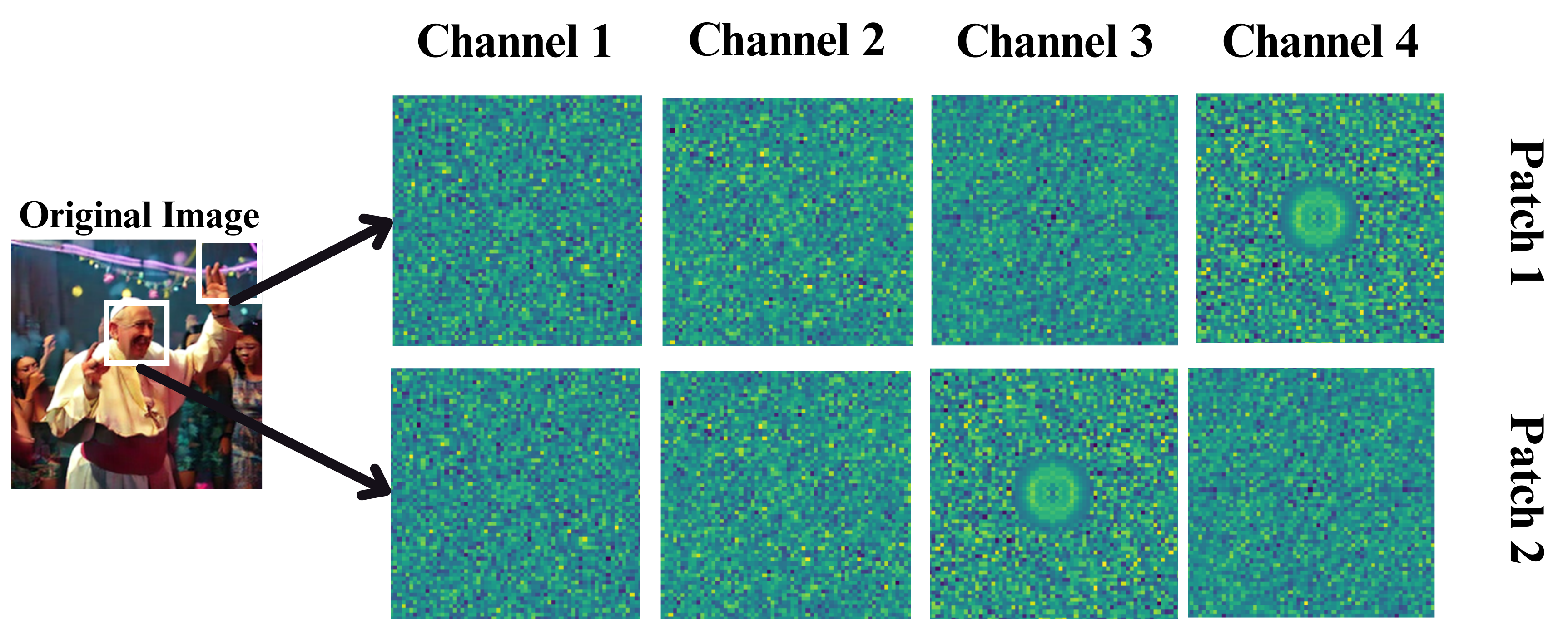}
    \vspace{-2mm}
    \caption{Illustration of multipatch watermarking, where watermark patterns are embedded in different channels of selected patches from the original image.}
    \label{fig:progression-multichannel-watermarking}
\end{figure}

\subsection{How to Watermark? - Strategies}
Once NMPs are identified, the next essential step is determining how to embed watermark signals within them. We experimented with four strategies: (i) Baseline watermarking with SoTA methods like ZoDiac \citep{zodiac} etc, (ii) watermark strength, (iii) single-channel strength watermarking, and (iv) multi-channel watermarking strength. Our results showed that multi-channel strength watermarking was the most effective approach. 

\textbf{Baseline watermarking with SoTA methods:} After identifying NMPs, our first step was to embed watermarks using state-of-the-art techniques like ZoDiac \citep{zodiac}, Stable Signature \citep{fernandez2023stablesignaturerootingwatermarks} and WAM \citep{sander2024watermarklocalizedmessages}. We are particularly interested in evaluating which method provides more resilient watermarking against visual paraphrase attacks.
\textbf{Watermark strength}: As argued by \citep{barman2024brittlenessaigeneratedimagewatermarking}, stronger paraphrasing removes watermarks more effectively. Therefore, using a higher watermark strength in these NMPs should make PECCAVI more resilient. 

Watermark strength is determined by the distance between rings within the watermark, with smaller distances indicating greater strength. For example, Channel 4 (Figure \ref{fig:progression-multichannel-watermarking}) shows a smaller ring distance (0.5), while Channel 3 reflects a larger distance (0.75). Strength values range from 0 to 1.0, depending on the number of paraphrases containing the NMP: $W_{s} = \max\left(0.1, 1 - 0.25 \cdot (n - 1)\right), \quad n \in \{1, 2, 3, 4, 5\}$. Here $n$ represent the number of regions that an NMP appears in out of the 5 paraphrases.
\textbf{Single-Channel Watermarking vs. Multi-Channel Watermarking}:
\textls[-7]{Fourier Space Watermarking \citep{gourrame2022fourier} embeds watermarks into the frequency domain of an image rather than the traditional spatial domain (pixels). This approach increases the watermark's resilience to image manipulations, as the image undergoes Fourier Transform decomposition before the watermark is added to its components. We can embed watermarks across same or different channels for each patch. Fig. \ref{fig:progression-multichannel-watermarking} provides an illustration adding watermarks in different channels.}

\textbf{Noisy Burnishing}:
Attackers may attempt to identify salient regions of an image to remove the watermark. This can be countered by adding adversarial noise to the watermarked image, which disrupts the detection of these salient regions, as proposed in \citep{jadena}. Fig. \ref{fig:noise_burnishing} demonstrates how salient regions become distorted following a noisy burnishing attack.

\begin{figure*}[ht!]
\centering
\includegraphics[width=1.0\linewidth]{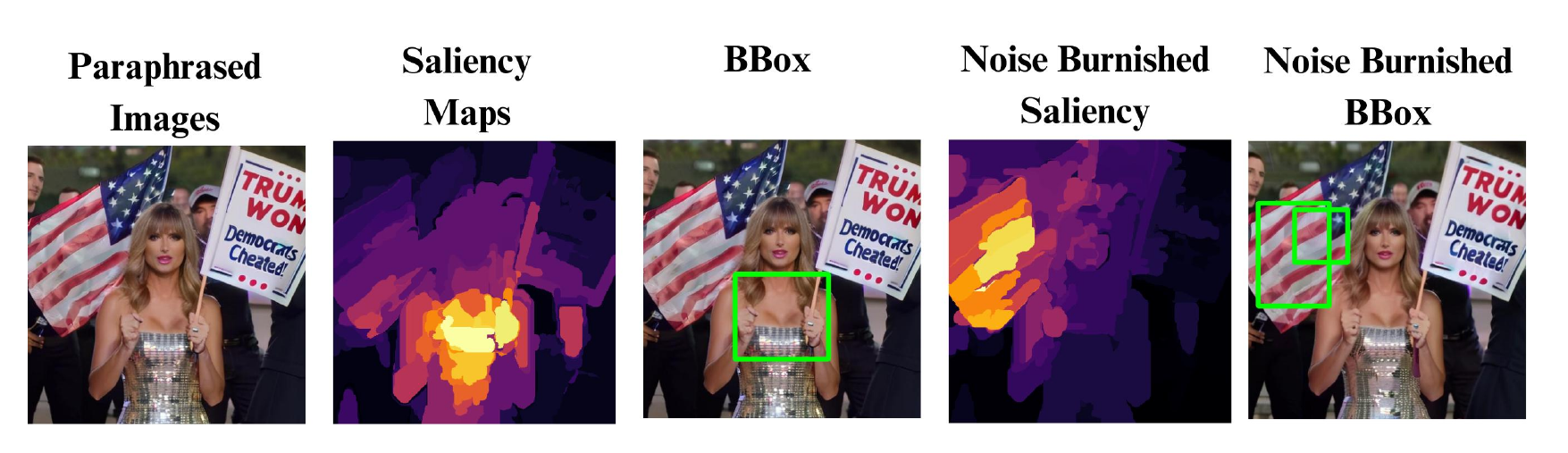}
\vspace{-2mm}
\caption{Noisy burnishing disrupts saliency detection in watermarked images, hindering attackers from locating NMPs or altering watermarked areas. This technique preserves the frequency-based watermark, ensuring high detectability while enhancing security against tampering.}
\label{fig:noise_burnishing}
\end{figure*}

\subsection{Paraphrase Attack Safety}
NMP-based watermarking faces two key challenges: (i) assessing the resilience of NMP-embedded watermarks against further paraphrasing, and (ii) anticipating potential countermeasures from attackers who may reverse-engineer methods to detect and distort NMPs, reducing watermark detectability. To address these, we propose two strategies: (a) random patching to embed additional watermarks, and (b) noisy burnishing to prevent NMP detection. Details of these techniques are provided below.
\textbf{Random Patching}: Since NMP detection relies on widely recognized saliency detection methods, such as XRAI \citep{9008576}, attackers could potentially reverse-engineer these methods to locate the salient regions where watermarks are embedded. To enhance security, we introduce a technique called random patching. This technique is simple yet effective: once all NMPs are detected and saved, we identify the smallest one among them and generate an additional NMP of the same shape at a random, non-overlapping location. The selection can be randomized using a vendor-specific pseudo-random algorithm. Watermarks are then embedded in these randomly placed patch, similar to the original patches, using either single-channel or multi-channel approaches.

\begin{figure*}[htbp!]
    \centering
    \resizebox{1.0\linewidth}{!}{
        \begin{tabular}{l}
            \begin{minipage}{\linewidth}\centering 
                \includegraphics[width=\linewidth]{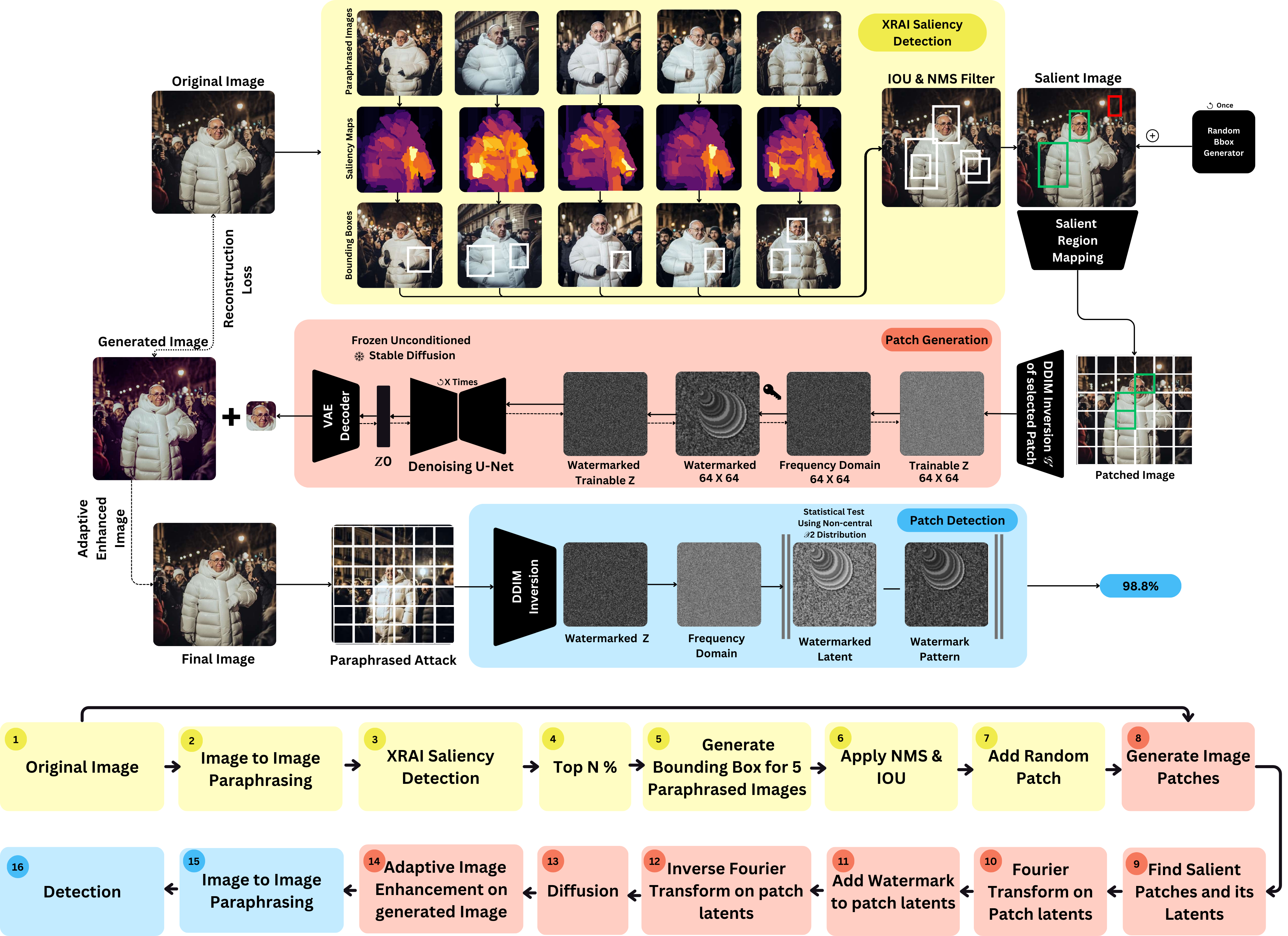}
            \end{minipage}
        \end{tabular}
    }
    \vspace{-3mm}
    \caption{The PECCAVI pipeline for image watermarking encompasses NMP detection, multi-channel watermark embedding, adaptive enhancement. These components collectively ensure robust, low-distortion watermarks that resist paraphrase attacks, safeguarding AI-generated images from unauthorized alterations.}
    \label{fig:peccavi_architecture}
    \vspace{-3mm}
\end{figure*}

\subsection{PECCAVI - Watermark Detection}  
The PECCAVI watermark detection process, illustrated in Figure \ref{fig:peccavi_architecture}, employs a brute-force approach to detect watermarks across all patches of an image. The highest detection score among all processed patches is selected as the final score. Additionally, the method scans multiple channels to identify watermark signals and ultimately computes a Watermark Detection Probability for the image.

\subsection{Adaptive Enhancement - Minimizing Distortion}
We apply adaptive image enhancement from \citep{zodiac} to improve watermarked image quality by blending it with the original:
$\bar{x}_0 = \hat{x}_0 + \gamma (x_0 - \hat{x}_0)
$ where $\gamma \in [0, 1]$ balances quality and watermark strength. The goal is to find the smallest $\gamma$ such that similarity $S(\bar{x}_0, x_0) \geq s^*$, typically using SSIM. An illustration is shown in the Figure \ref{fig:adaptive_enhancement}.

\vspace{-1mm}
\section{Efficacy of PECCAVI}  \label{sec:performance}

To assess PECCAVI’s robustness, we evaluate two key dimensions: distortion and detectability. For image quality distortion, we use metrics such as PSNR and SSIM to measure the watermark’s impact on visual fidelity, both perceptually and structurally. For watermark detectability, we analyze PECCAVI’s resilience against classical attacks like brightness adjustments, Gaussian noise, JPEG compression, and varying paraphrasing strengths using the Average WDP. A summary of these metrics, presented in Table \ref{tab:results}, highlights PECCAVI’s effectiveness in preserving high image quality while ensuring robust watermark retention under diverse attack scenarios.

Meta's Watermark Anything Model (WAM) \citep{sander2024watermarklocalizedmessages} (\emph{released on 11\textsuperscript{th} Nov}) enables imperceptible, localized image watermarking, embedding, locating, and decoding multiple watermarks in small regions of high-resolution images. Our evaluation shows that PECCAVI outperforms WAM in resisting visual paraphrasing attacks.

\subsection{Choice of T2I models} \label{sec:llm}

To evaluate the efficacy of PECCAVI, we tested it across diverse T2I models, including Stable Diffusion 3 (SD 3) \citep{esser2024scalingrectifiedflowtransformers,stabilityai2024stable_diffusion_3}, Stable Diffusion XL (SDXL) \citep{podell2023sdxlimprovinglatentdiffusion}, Stable Diffusion 2.1 (SD 2.1) \citep{Rombach_2022_CVPR,stabilityai2023stable_diffusion_2_1}, DALL-E 3 \citep{BetkerImprovingIG}, and Midjourney 6 \citep{Midjourney2024}. This process produced a dataset we call MS COCO\textsubscript{AI}, where captions and images from the original MS COCO dataset \citep{Lin2014MicrosoftCC} were fed into these models to generate and store corresponding images. A snapshot of the data can be viewed \href{https://huggingface.co/datasets/anonymous1233/COCO_AI}{here}. Results present in Table \ref{tab:results} present an average over all the images in MS COCO\textsubscript{AI}.

\subsection{Measuring Paraphrase Attack Safety}

Paraphrased images were generated at different strengths \( s \), with lower \( s \)-values keeping more original details and higher values allowing greater alteration. WDP assesses watermark retention, while SSIM measures similarity to the original image. PECCAVI shows high WDP at lower strengths, retaining watermark integrity even through moderate paraphrasing. Detection gradually decreases with higher \( s \) but remains effective, as shown in Table \ref{tab:results}.

\subsection{Measuring Distortion}

A key requirement in image watermarking is minimal distortion of the original content. We assess this distortion using metrics like PSNR \citep{8980171} and SSIM \citep{ssim}. Further details are reported in the Appendix: Distortion vs. Detectability. 
Together, these metrics provide a balanced view of pixel-level (PSNR, SSIM) distortion, helping us assess overall image quality. For distortion metric results refer to Table \ref{tab:results}.

\subsection{Results}
We compare the performance of PECCAVI watermarking scheme with various post-process image watermarking methods such as \citep{zodiac}, \citep{2019stegastamp}, \citep{fernandez2023stablesignaturerootingwatermarks}, \citep{navas2008dwt} and \citep{cin_watermark}. The methods were compared under the following attack schemes: (i) Brightness Enhancement with a factor of 0.5, (ii) Gaussian Noise with a std of 0.05, (iii) JPEG compression with a quality factor of 50, and (iv) Visual Paraphrasing \citep{barman2024brittlenessaigeneratedimagewatermarking}, using \texttt{stable-diffusion-xl-base-1.0} with image captions and paraphrase strengths of 0.1 and 0.2. We further test our method on VAE-based image compression model \citep{ballé2018variationalimagecompressionscale} and \citep{cheng2020} with a quality setting of 3, Stable Diffusion-based image regeneration model  \citep{zhao2024invisibleimagewatermarksprovably} with 60 denoising steps using stable-diffusion-2-1-base; results provided in supplementary material.

The results were produced on 100 images randomly sampled from the COCO Dataset \citep{Lin2014MicrosoftCC}.

\newcommand{\gradient}[3]{#1}
\begin{table*}[ht]
    \centering
    \resizebox{\textwidth}{!}{%
    \begin{tabular}{l|c|cc|c|ccc|cc}
    \midrule
    & & \multicolumn{2}{c|}{Image Quality} & \multicolumn{6}{c}{Avg. Watermark Detection Probability (WDP)} \\
    \addlinespace[2pt]
    \cline{3-10}
    \addlinespace[2pt]
    & & & & \multicolumn{1}{c|}{Pre-Attack} & \multicolumn{5}{c}{Post-Attack} \\
    \addlinespace[2pt]
    \cline{5-10}
    \addlinespace[2pt]
    Method & $\lambda$ & PSNR & SSIM & & Brightness & Gaussian Noise & JPEG & Paraphrase (s=0.1) & Paraphrase (s=0.2) \\
    \midrule
    DwtDctSVD & - & \gradient{41.04}{28}{47} & \gradient{0.988}{0}{1} & \gradient{0.98}{0}{1} & \gradient{0.01}{0}{1} & \gradient{0.14}{0}{1} & \gradient{0.65}{0}{1} & \gradient{0.00}{0}{1} & \gradient{0.00}{0}{1} \\
    Stable Signature& -   & 42.91 & 0.98  & 0.99           & 0.75         & 0.73         & 0.65         & 0.59         & 0.51         \\
    WAM             & -   & 46.05 & 1.00  & \cellcolor{green!20}1.00  & 0.62         & 0.61         & 0.58         & 0.63         & 0.56         \\
    ZoDiac          & -   & 28.47 & 0.92  & \cellcolor{green!20}1.00  & \cellcolor{green!20}0.92 & \cellcolor{green!20}0.90 & \cellcolor{green!20}0.89 & \cellcolor{green!20}0.81 & \cellcolor{green!20}0.70 \\
    \midrule
    \multicolumn{10}{l}{\textbf{PECCAVI with different saliency methods}} \\
    \midrule
                     & Top 30 & 31.50 & 0.95 & 0.96           & 0.96         & 0.95         & 0.96         & 0.72         & 0.69         \\
    PECCAVI (Vanilla Integrated) & Top 40 & 31.26 & 0.94 & \cellcolor{green!20}0.95  & \cellcolor{green!20}0.97 & \cellcolor{green!20}0.97 & \cellcolor{green!20}0.95 & 0.72         & 0.68         \\
                     & Top 50 & 31.31 & 0.95 & 0.96           & 0.96         & 0.97         & 0.97         & 0.73         & 0.68         \\
    \addlinespace[2pt]
    \hdashline
    \addlinespace[2pt]
                     & Top 30 & 30.64 & 0.94 & \cellcolor{green!20}0.98  & \cellcolor{green!20}0.94 & \cellcolor{green!20}0.95 & \cellcolor{green!20}0.95 & \cellcolor{green!20}0.84 & \cellcolor{green!20}0.79 \\
    PECCAVI (MSI Net)       & Top 40 & 30.57 & 0.94 & \cellcolor{green!20}0.99  & \cellcolor{green!20}0.98 & \cellcolor{green!20}0.98 & \cellcolor{green!20}0.97 & \cellcolor{green!20}0.83 & \cellcolor{green!20}0.80 \\
                     & Top 50 & 30.71 & 0.94 & \cellcolor{green!20}0.99  & \cellcolor{green!20}0.98 & \cellcolor{green!20}0.98 & \cellcolor{green!20}0.98 & \cellcolor{green!20}0.87 & \cellcolor{green!20}0.83 \\
    \addlinespace[2pt]
    \hdashline
    \addlinespace[2pt]
                     & Top 30 & 29.56 & 0.93 & \cellcolor{green!20}0.99  & \cellcolor{green!20}0.98 & \cellcolor{green!20}0.99 & \cellcolor{green!20}0.99 & \cellcolor{green!20}0.92 & \cellcolor{green!20}0.87 \\
    PECCAVI (XRAI)          & Top 40 & 29.87 & 0.93 & \cellcolor{green!20}0.99  & \cellcolor{green!20}0.99 & \cellcolor{green!20}0.99 & \cellcolor{green!20}0.98 & \cellcolor{green!20}0.90 & \cellcolor{green!20}0.84 \\
                     & Top 50 & 29.84 & 0.93 & \cellcolor{green!20}0.99  & \cellcolor{green!20}0.99 & \cellcolor{green!20}0.99 & \cellcolor{green!20}0.99 & \cellcolor{green!20}0.90 & \cellcolor{green!20}0.85 \\
    \hline
    \end{tabular}
     }
    \caption{Watermarked image quality is compared in terms of PSNR and SSIM scores. Watermark robustness is compared based on Average WDP before and after attacks on the MS-COCO dataset.}
    \label{tab:results}
    \vspace{-3mm}
\end{table*}

\begin{figure}
    \centering
    \includegraphics[width=1\linewidth]{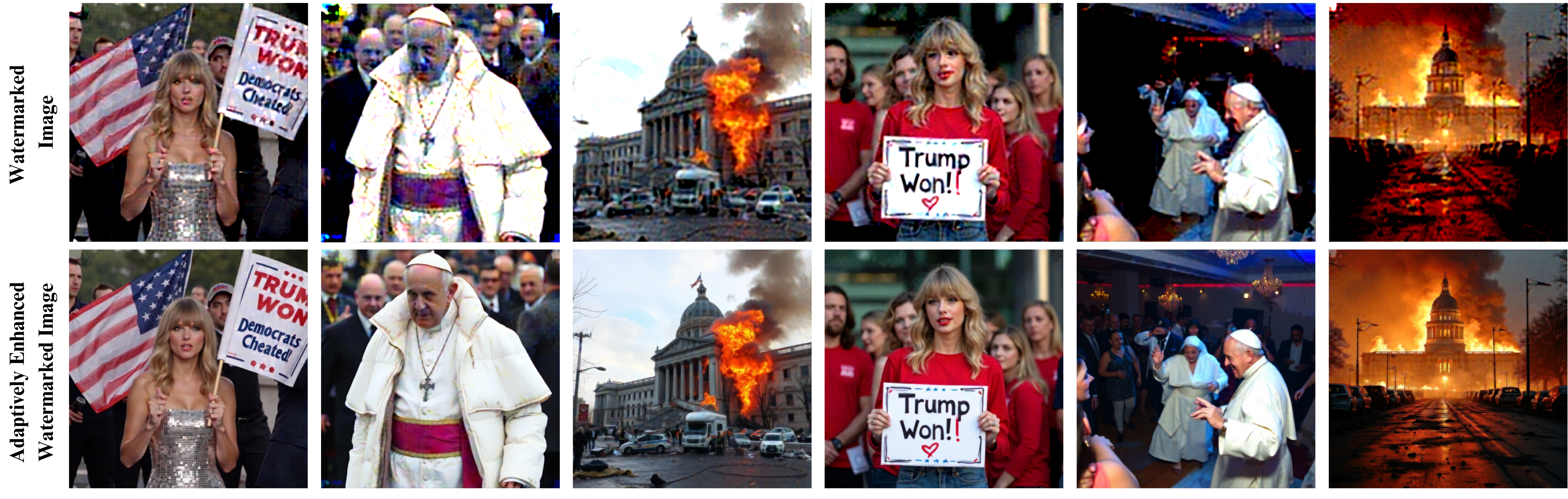}
    \caption{Comparison of two sets of images before and after
adaptive enhancement, which optimizes watermark detectability while minimizing visual distortion.}
    \label{fig:Adaptive Enhancement}
\end{figure}

\begin{figure}
    \centering
    \includegraphics[width=1\linewidth]{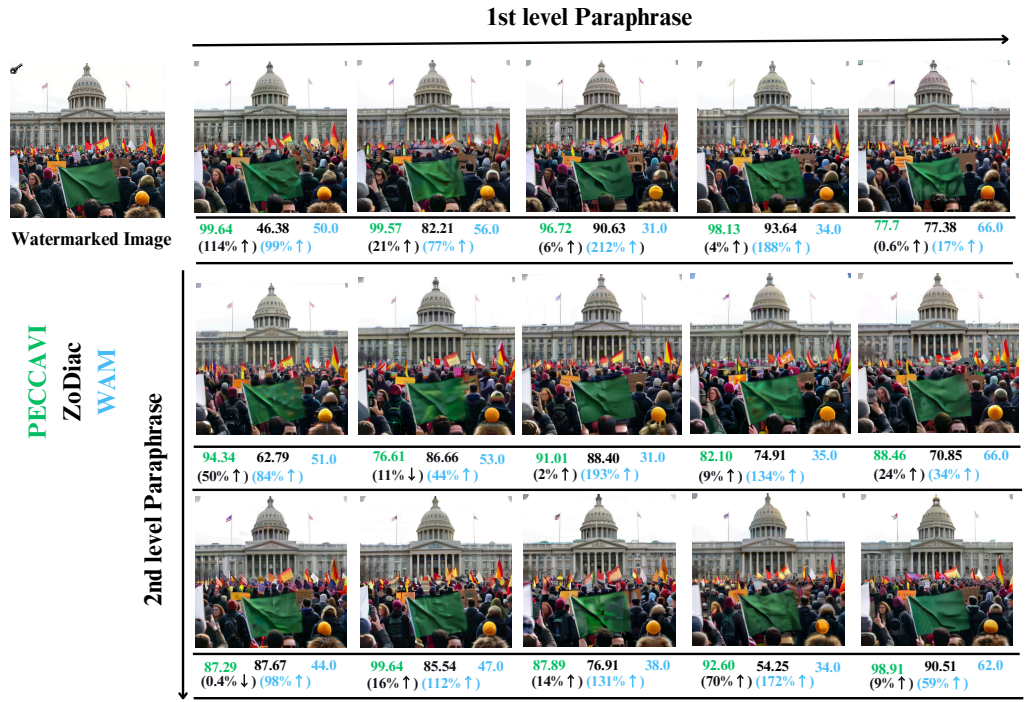}
    \caption{The cumulative impact of successive paraphrasing attacks on PECCAVI (green) and ZoDiac (black) watermarked images is depicted, with detection scores. PECCAVI shows superior resilience, maintaining stable scores under high-strength paraphrasing attacks, demonstrating its durability over ZoDiac.}
    \label{fig:Distortion}
\end{figure}

\vspace{-1mm}
\section{Conclusion}
\vspace{-1mm}
This paper introduces PECCAVI is the first visual paraphrase attack-safe, distortion-free image watermarking technique. With the rise of AI-generated misinformation, we believe PECCAVI will contribute significantly to the greater social good. It surpasses existing watermarking techniques like ZoDiac and WAM in performance, though it requires substantial computational resources. Due to space constraints, comparative analysis is provided in the appendix.

{
    \small
    \bibliographystyle{ieeenat_fullname}
    \bibliography{main}

\begin{thebibliography}{49}
\providecommand{\natexlab}[1]{#1}
\providecommand{\url}[1]{\texttt{#1}}
\expandafter\ifx\csname urlstyle\endcsname\relax
  \providecommand{\doi}[1]{doi: #1}\else
  \providecommand{\doi}{doi: \begingroup \urlstyle{rm}\Url}\fi

\bibitem[AI(2023)]{stabilityai2023stable_diffusion_2_1}
Stability AI.
\newblock Stable diffusion 2.1, 2023.
\newblock Available at: \url{https://huggingface.co/stabilityai/stable-diffusion-2-1}.

\bibitem[AI(2024)]{stabilityai2024stable_diffusion_3}
Stability AI.
\newblock Stable diffusion 3: Research paper.
\newblock 2024.

\bibitem[Baig et~al.(2019)Baig, Moinuddin, and Khan]{8980171}
Md~Amir Baig, Athar~A. Moinuddin, and E. Khan.
\newblock Psnr of highest distortion region: An effective image quality assessment method.
\newblock In \emph{2019 International Conference on Electrical, Electronics and Computer Engineering (UPCON)}, pages 1--4, 2019.

\bibitem[Ballé et~al.(2018)Ballé, Minnen, Singh, Hwang, and Johnston]{ballé2018variationalimagecompressionscale}
Johannes Ballé, David Minnen, Saurabh Singh, Sung~Jin Hwang, and Nick Johnston.
\newblock Variational image compression with a scale hyperprior, 2018.

\bibitem[Barman et~al.(2024)Barman, Sharma, Aziz, Bajpai, Biswas, Sharma, Jain, Chadha, Sheth, and Das]{barman2024brittlenessaigeneratedimagewatermarking}
Niyar~R Barman, Krish Sharma, Ashhar Aziz, Shashwat Bajpai, Shwetangshu Biswas, Vasu Sharma, Vinija Jain, Aman Chadha, Amit Sheth, and Amitava Das.
\newblock The brittleness of ai-generated image watermarking techniques: Examining their robustness against visual paraphrasing attacks, 2024.

\bibitem[Betker et~al.()Betker, Goh, Jing, TimBrooks, Wang, Li, LongOuyang, JuntangZhuang, JoyceLee, YufeiGuo, WesamManassra, PrafullaDhariwal, CaseyChu, YunxinJiao, and Ramesh]{BetkerImprovingIG}
James Betker, Gabriel Goh, Li Jing, † TimBrooks, Jianfeng Wang, Linjie Li, † LongOuyang, † JuntangZhuang, † JoyceLee, † YufeiGuo, † WesamManassra, † PrafullaDhariwal, † CaseyChu, † YunxinJiao, and Aditya Ramesh.
\newblock Improving image generation with better captions.

\bibitem[{Business Standard}(2024)]{businessstandard2024}
{Business Standard}.
\newblock Openai mulls watermarking chatgpt generated text, but treads with caution.
\newblock \emph{Business Standard}, 2024.

\bibitem[california legislature(2023)]{california_act}
california legislature.
\newblock Ab-3211 california digital content provenance standards.
\newblock 2023.

\bibitem[Chakraborty et~al.(2023)Chakraborty, Tonmoy, Zaman, Gautam, Kumar, Sharma, Barman, Gupta, Jain, Chadha, Sheth, and Das]{chakraborty-etal-2023-counter}
Megha Chakraborty, S.M Towhidul~Islam Tonmoy, S~M~Mehedi Zaman, Shreya Gautam, Tanay Kumar, Krish Sharma, Niyar Barman, Chandan Gupta, Vinija Jain, Aman Chadha, Amit Sheth, and Amitava Das.
\newblock Counter {T}uring test ({CT}2): {AI}-generated text detection is not as easy as you may think - introducing {AI} detectability index ({ADI}).
\newblock In \emph{Proceedings of the 2023 Conference on Empirical Methods in Natural Language Processing}, pages 2206--2239, Singapore, 2023. Association for Computational Linguistics.

\bibitem[Cheng et~al.(2020)Cheng, Sun, Takeuchi, and Katto]{cheng2020}
Zhengxue Cheng, Heming Sun, Masaru Takeuchi, and Jiro Katto.
\newblock Learned image compression with discretized gaussian mixture likelihoods and attention modules, 2020.

\bibitem[CNBC(2024)]{gates_misinformation}
CNBC.
\newblock Bill gates says this is the no. 1 unsolvable problem facing today’s young people: ‘the harm is done’.
\newblock 2024.

\bibitem[Commission(2022)]{EU_code_of_conduct_2022}
European Commission.
\newblock Eu code of conduct against online hate speech: latest evaluation shows slowdown in progress.
\newblock 2022.

\bibitem[Cox et~al.(1997)Cox, Kilian, Leighton, and Shamoon]{cox1997secure}
Ingemar~J Cox, Joe Kilian, F~Thomson Leighton, and Talal Shamoon.
\newblock Secure spread spectrum watermarking for multimedia.
\newblock \emph{IEEE Transactions on Image Processing}, 6\penalty0 (12):\penalty0 1673--1687, 1997.

\bibitem[Esser et~al.(2024)Esser, Kulal, Blattmann, Entezari, Müller, Saini, Levi, Lorenz, Sauer, Boesel, Podell, Dockhorn, English, Lacey, Goodwin, Marek, and Rombach]{esser2024scalingrectifiedflowtransformers}
Patrick Esser, Sumith Kulal, Andreas Blattmann, Rahim Entezari, Jonas Müller, Harry Saini, Yam Levi, Dominik Lorenz, Axel Sauer, Frederic Boesel, Dustin Podell, Tim Dockhorn, Zion English, Kyle Lacey, Alex Goodwin, Yannik Marek, and Robin Rombach.
\newblock Scaling rectified flow transformers for high-resolution image synthesis, 2024.

\bibitem[EUROPOL(2022)]{EUROPOL}
EUROPOL.
\newblock Facing reality?: Law enforcement and the challenge of deepfakes.
\newblock 2022.

\bibitem[Fernandez et~al.(2023)Fernandez, Couairon, Jégou, Douze, and Furon]{fernandez2023stablesignaturerootingwatermarks}
Pierre Fernandez, Guillaume Couairon, Hervé Jégou, Matthijs Douze, and Teddy Furon.
\newblock The stable signature: Rooting watermarks in latent diffusion models, 2023.

\bibitem[{Future of Life Institute}(2023)]{aihalt2023}
{Future of Life Institute}.
\newblock Pause giant ai experiments: An open letter, 2023.

\bibitem[Gao et~al.(2022)Gao, Guo, Juefei-Xu, Yu, Fu, Feng, Liu, and Wang]{jadena}
Ruijun Gao, Qing Guo, Felix Juefei-Xu, Hongkai Yu, Huazhu Fu, Wei Feng, Yang Liu, and Song Wang.
\newblock Can you spot the chameleon? adversarially camouflaging images from co-salient object detection.
\newblock In \emph{Proceedings of the IEEE/CVF Conference on Computer Vision and Pattern Recognition}, pages 2150--2159, 2022.

\bibitem[Gourrame et~al.(2022)Gourrame, Ros, Douzi, Harba, and Riad]{gourrame2022fourier}
Khadija Gourrame, Frederic Ros, Hassan Douzi, Rachid Harba, and Rabia Riad.
\newblock Fourier image watermarking: Print-cam application.
\newblock \emph{Electronics}, 11\penalty0 (2):\penalty0 266, 2022.

\bibitem[Guardian(2023)]{hinton_misinformation}
The Guardian.
\newblock Godfather of ai’ geoffrey hinton quits google and warns over dangers of misinformation.
\newblock 2023.

\bibitem[Janjeva et~al.(2023)Janjeva, Harris, Mercer, Kasprzyk, and Gausen]{janjevagenai2023}
Ardi Janjeva, Alexander Harris, Sarah Mercer, Alexander Kasprzyk, and Anna Gausen.
\newblock The rapid rise of generative ai: Assessing risks to safety and security, 2023.

\bibitem[Kapishnikov et~al.(2019)Kapishnikov, Bolukbasi, Viegas, and Terry]{9008576}
Andrei Kapishnikov, Tolga Bolukbasi, Fernanda Viegas, and Michael Terry.
\newblock Xrai: Better attributions through regions.
\newblock In \emph{2019 IEEE/CVF International Conference on Computer Vision (ICCV)}, pages 4947--4956, 2019.

\bibitem[Kharpal(2018)]{pichai_misinformation}
Arjun Kharpal.
\newblock Google ceo sundar pichai: Fears about artificial intelligence are very legitimate, he says in post interview.
\newblock \emph{The Washington Post}, 2018.

\bibitem[Kirchenbauer et~al.(2023)Kirchenbauer, Geiping, Wen, Katz, Miers, and Goldstein]{Kirchenbauer2023AWF}
John Kirchenbauer, Jonas Geiping, Yuxin Wen, Jonathan Katz, Ian Miers, and Tom Goldstein.
\newblock A watermark for large language models.
\newblock In \emph{International Conference on Machine Learning}, 2023.

\bibitem[Kroner et~al.(2020)Kroner, Senden, Driessens, and Goebel]{kroner2020contextual}
Alexander Kroner, Mario Senden, Kurt Driessens, and Rainer Goebel.
\newblock Contextual encoder-decoder network for visual saliency prediction.
\newblock \emph{Neural Networks}, 129:\penalty0 261--270, 2020.

\bibitem[Lin et~al.(2014)Lin, Maire, Belongie, Hays, Perona, Ramanan, Doll{\'a}r, and Zitnick]{Lin2014MicrosoftCC}
Tsung-Yi Lin, Michael Maire, Serge~J. Belongie, James Hays, Pietro Perona, Deva Ramanan, Piotr Doll{\'a}r, and C.~Lawrence Zitnick.
\newblock Microsoft coco: Common objects in context.
\newblock In \emph{European Conference on Computer Vision}, 2014.

\bibitem[Loch(2010)]{brightness_adjustment}
Francis~G. Loch.
\newblock Image processing algorithms part 4: Brightness adjustment.
\newblock \emph{The Crypt Mag}, 2010.

\bibitem[Ma et~al.(2022)Ma, Guo, Hou, Yang, Li, Jia, and Xie]{cin_watermark}
Rui Ma, Mengxi Guo, Yi Hou, Fan Yang, Yuan Li, Huizhu Jia, and Xiaodong Xie.
\newblock Towards blind watermarking: Combining invertible and non-invertible mechanisms.
\newblock In \emph{Proceedings of the 30th ACM International Conference on Multimedia}, pages 1532--1542, 2022.

\bibitem[Midjourney(2024)]{Midjourney2024}
Midjourney.
\newblock Midjourney: A new era of ai-generated art, 2024.

\bibitem[Navas et~al.(2008)]{navas2008dwt}
A. Navas et~al.
\newblock Digital watermarking techniques: A review.
\newblock \emph{Journal of Information Technology}, 23\penalty0 (4):\penalty0 345--360, 2008.

\bibitem[Nie et~al.(2022)Nie, Guo, Huang, Xiao, Vahdat, and Anandkumar]{pmlr-v162-nie22a}
Weili Nie, Brandon Guo, Yujia Huang, Chaowei Xiao, Arash Vahdat, and Animashree Anandkumar.
\newblock Diffusion models for adversarial purification.
\newblock In \emph{Proceedings of the 39th International Conference on Machine Learning}, pages 16805--16827. PMLR, 2022.

\bibitem[Podell et~al.(2023)Podell, English, Lacey, Blattmann, Dockhorn, Müller, Penna, and Rombach]{podell2023sdxlimprovinglatentdiffusion}
Dustin Podell, Zion English, Kyle Lacey, Andreas Blattmann, Tim Dockhorn, Jonas Müller, Joe Penna, and Robin Rombach.
\newblock Sdxl: Improving latent diffusion models for high-resolution image synthesis, 2023.

\bibitem[Podilchuk and Ramchandran(1998)]{podilchuk1998image}
C. Podilchuk and K. Ramchandran.
\newblock Image watermarking: Capacity issues and applications.
\newblock In \emph{Proceedings of the International Conference on Image Processing (ICIP)}, pages 445--448. IEEE, 1998.

\bibitem[{PyTorch Discussion Forum}(2022)]{pytorch_max_seed_discussion}
{PyTorch Discussion Forum}.
\newblock What is the max seed you can set up?, 2022.
\newblock Accessed: 2024-11-13.

\bibitem[Ramesh et~al.(2021)Ramesh, Pavlov, Goh, Gray, Voss, Radford, Chen, and Sutskever]{ramesh2021zeroshottexttoimagegeneration}
Aditya Ramesh, Mikhail Pavlov, Gabriel Goh, Scott Gray, Chelsea Voss, Alec Radford, Mark Chen, and Ilya Sutskever.
\newblock Zero-shot text-to-image generation, 2021.

\bibitem[Ramesh et~al.(2022)Ramesh, Dhariwal, Nichol, Chu, and Chen]{ramesh2022hierarchicaltextconditionalimagegeneration}
Aditya Ramesh, Prafulla Dhariwal, Alex Nichol, Casey Chu, and Mark Chen.
\newblock Hierarchical text-conditional image generation with clip latents, 2022.

\bibitem[Rombach et~al.(2022)Rombach, Blattmann, Lorenz, Esser, and Ommer]{Rombach_2022_CVPR}
Robin Rombach, Andreas Blattmann, Dominik Lorenz, Patrick Esser, and Bj\"orn Ommer.
\newblock High-resolution image synthesis with latent diffusion models.
\newblock In \emph{Proceedings of the IEEE/CVF Conference on Computer Vision and Pattern Recognition (CVPR)}, pages 10684--10695, 2022.

\bibitem[Sadasivan et~al.(2024)Sadasivan, Kumar, Balasubramanian, Wang, and Feizi]{sadasivan2024aigeneratedtextreliablydetected}
Vinu~Sankar Sadasivan, Aounon Kumar, Sriram Balasubramanian, Wenxiao Wang, and Soheil Feizi.
\newblock Can ai-generated text be reliably detected?, 2024.

\bibitem[Saharia et~al.(2022)Saharia, Chan, Saxena, Li, Whang, Denton, Ghasemipour, Ayan, Mahdavi, Lopes, Salimans, Ho, Fleet, and Norouzi]{saharia2022photorealistictexttoimagediffusionmodels}
Chitwan Saharia, William Chan, Saurabh Saxena, Lala Li, Jay Whang, Emily Denton, Seyed Kamyar~Seyed Ghasemipour, Burcu~Karagol Ayan, S.~Sara Mahdavi, Rapha~Gontijo Lopes, Tim Salimans, Jonathan Ho, David~J Fleet, and Mohammad Norouzi.
\newblock Photorealistic text-to-image diffusion models with deep language understanding, 2022.

\bibitem[Sander et~al.(2024)Sander, Fernandez, Durmus, Furon, and Douze]{sander2024watermarklocalizedmessages}
Tom Sander, Pierre Fernandez, Alain Durmus, Teddy Furon, and Matthijs Douze.
\newblock Watermark anything with localized messages, 2024.

\bibitem[Tancik et~al.(2020)Tancik, Mildenhall, and Ng]{2019stegastamp}
Matthew Tancik, Ben Mildenhall, and Ren Ng.
\newblock Stegastamp: Invisible hyperlinks in physical photographs.
\newblock In \emph{IEEE Conference on Computer Vision and Pattern Recognition (CVPR)}, 2020.

\bibitem[Thomson et~al.(2020)Thomson, Angus, and Dootson]{Thomson2020}
T.J. Thomson, Daniel Angus, and Paula Dootson.
\newblock 3.2 billion images and 720,000 hours of video are shared online daily: Can you sort real from fake?
\newblock \emph{The Conversation}, 2020.
\newblock Accessed: 2024-11-04.

\bibitem[Wang et~al.(2004)Wang, Bovik, Sheikh, and Simoncelli]{ssim}
Zhou Wang, Alan~C Bovik, Hamid~R Sheikh, and Eero~P Simoncelli.
\newblock Image quality assessment: from error visibility to structural similarity.
\newblock \emph{IEEE transactions on image processing}, 13\penalty0 (4):\penalty0 600--612, 2004.

\bibitem[Wen et~al.(2023)Wen, Kirchenbauer, Geiping, and Goldstein]{wen2023treerings}
Yuxin Wen, John Kirchenbauer, Jonas Geiping, and Tom Goldstein.
\newblock Tree-rings watermarks: Invisible fingerprints for diffusion images.
\newblock In \emph{Thirty-seventh Conference on Neural Information Processing Systems}, 2023.

\bibitem[Yadav et~al.(2012)]{jpeg_compression}
Rajesh~K. Yadav et~al.
\newblock Study and analysis of wavelet based image compression techniques.
\newblock \emph{International Journal of Engineering, Science and Technology}, 4\penalty0 (1):\penalty0 1--7, 2012.

\bibitem[Zhang et~al.(2024)Zhang, Liu, Martin, Bearfield, Brun, and Guan]{zodiac}
Lijun Zhang, Xiao Liu, Antoni~Viros Martin, Cindy~Xiong Bearfield, Yuriy Brun, and Hui Guan.
\newblock Attack-resilient image watermarking using stable diffusion, 2024.

\bibitem[Zhao et~al.(2023)Zhao, Zhang, Wang, and Li]{zhao2023invisible}
Xuandong Zhao, Kexun Zhang, Yu-Xiang Wang, and Lei Li.
\newblock Generative autoencoders as watermark attackers: Analyses of vulnerabilities and threats.
\newblock \emph{arXiv preprint arXiv:2306.01953}, 2023.

\bibitem[Zhao et~al.(2024)Zhao, Zhang, Su, Vasan, Grishchenko, Kruegel, Vigna, Wang, and Li]{zhao2024invisibleimagewatermarksprovably}
Xuandong Zhao, Kexun Zhang, Zihao Su, Saastha Vasan, Ilya Grishchenko, Christopher Kruegel, Giovanni Vigna, Yu-Xiang Wang, and Lei Li.
\newblock Invisible image watermarks are provably removable using generative ai, 2024.

\bibitem[Zhou et~al.(2024)Zhou, Zhang, Song, Zheng, Lu, Liu, Chen, and Xi]{zhou2024zodiaccardiologistlevelllmframework}
Yuan Zhou, Peng Zhang, Mengya Song, Alice Zheng, Yiwen Lu, Zhiheng Liu, Yong Chen, and Zhaohan Xi.
\newblock Zodiac: A cardiologist-level llm framework for multi-agent diagnostics, 2024.

\end{thebibliography}
}
\newpage

\end{document}